%% file: main.tex
\documentclass{article}
\usepackage{fullpage}
\title{PRIMA: Planner-Reasoner Inside a Multi-task Reasoning Agent}
\author{
  Daoming Lyu\\
  Auburn University, Auburn, AL\\
  \texttt{daoming.lyu@auburn.edu} \\
  \and
  Bo Liu\\
  Auburn University, Auburn, AL\\
  \texttt{boliu@auburn.edu} \\
  \and
  Jianshu Chen\\
  Tencent AI Lab, Bellevue, WA\\
  \texttt{jianshuchen@tencent.com} \\
}

\date{}

\usepackage[utf8]{inputenc} 
\usepackage[T1]{fontenc}    
\usepackage{hyperref}       
\usepackage{url}            
\usepackage{booktabs}       
\usepackage{amsfonts}       
\usepackage{nicefrac}       
\usepackage{microtype}      
\usepackage{xcolor}         
\usepackage{natbib}

\input{math_commands.tex}

\usepackage{threeparttable}
\usepackage{multirow,multicol}
\usepackage{wrapfig}
\usepackage{microtype}
\usepackage{graphicx}
\usepackage{subcaption}
\usepackage{booktabs} 
\usepackage{graphicx}
\usepackage{hyperref}
\usepackage{url}
\usepackage{color}
\usepackage{hyperref}

\newcommand{\state}{s}

\newcommand{\action}{a}

\begin{document}
  
\maketitle

\begin{abstract}
We consider the problem of multi-task reasoning (MTR), where an agent can solve multiple tasks via (first-order) logic reasoning. This capability is essential for human-like intelligence due to its strong generalizability and simplicity for handling multiple tasks. However, a major challenge in developing effective MTR is the intrinsic conflict between reasoning capability and efficiency. An MTR-capable agent must master a large set of ``skills'' to tackle diverse tasks, but executing a particular task at the inference stage requires only a small subset of immediately relevant skills. How can we maintain broad reasoning capability and also efficient specific-task performance? To address this problem, we propose a Planner-Reasoner framework capable of state-of-the-art MTR capability and high efficiency. The Reasoner models shareable (first-order) logic deduction rules, from which the Planner selects a subset to compose into efficient reasoning paths. The entire model is trained in an end-to-end manner using deep reinforcement learning, and experimental studies over a variety of domains validate its effectiveness.
\end{abstract}

\section{Introduction}
\label{sec:intro}
Multi-task learning (MTL)~\citep{zhang2021survey,zhou2011malsar} demonstrates superior sample complexity and generalizability compared with the conventional ``one model per task'' style to solve multiple tasks.
Recent research has additionally leveraged the great success of deep learning~\citep{lecun2015deep} to empower learning deep multi-task models~\citep{zhang2021survey,crawshaw2020multi}. Deep MTL models either learn a common multi-task feature representation by sharing several bottom layers of deep neural networks~\citep{zllt14,lmzcl15,zlzskyj15,mstgsvwy15,llc15}, or learn task-invariant and task-specific neural modules~\citep{shinohara16,lqh17} via generative adversarial networks~\citep{goodfellow2014generative}.
Although MTL is successful in many applications, a major challenge is the often impractically large MTL models. Although still smaller than piling up all models across different tasks, existing MTL models are significantly larger than a single model for tackling a specific task.
This results from the intrinsic conflict underlying all MTL algorithms: balancing \textit{across-task generalization capability} to perform different tasks with \textit{single-task efficiency} in executing a specific task. 
On the one hand, good generalization ability requires an MTL agent to be equipped with a large set of skills that can be combined to solve many different tasks. On the other hand, solving one particular task does not require all these skills. Instead, the agent needs to compose only a (small) subset of these skills into an efficient solution for a specific task. This conflict often hobbles existing MTL approaches.

This paper focuses on multi-task reasoning (MTR), a subarea of MTL that uses logic reasoning to solve multiple tasks. MTR is ubiquitous in human reasoning, where humans construct different reasoning paths for multiple tasks from the \emph{same} set of reasoning skills.
Conventional deep learning, although capable of strong expressive power, falls short in reasoning capabilities~\citep{bengio2019from}. Considerable research has been devoted to endowing deep learning with logic reasoning abilities, the results of which include Deep Neural Reasoning~\citep{jaeger2016deep}, Neural Logic Reasoning~\citep{nsai:besold2017neural,nsai:bader2004integration,nsai:bader2005dimensions}, Neural Logic Machines~\citep{nlm}, and other approaches~\citep{nsai:besold2017neural,nsai:bader2004integration,nsai:bader2005dimensions}. However, these approaches consider only single-task reasoning rather than a multi-task setting, and applying existing MTL approaches to learning these neural reasoning models leads to the same conflict between \emph{across-task generalization} and \emph{single-task efficiency}.

To strike a balance between reasoning capability and efficiency in MTR, we develop a \textit{\textbf{P}lanner-\textbf{R}easoner architecture \textbf{I}nside a \textbf{M}ulti-task reasoning \textbf{A}gent} (PRIMA) (Section \ref{sec:PRIMA}), wherein the reasoner defines a set of neural logic operators for modeling reusable reasoning meta-rules (``skills'') across tasks (Section \ref{sec:PRIMA:reasoner}). 
When defining the logic operators, we focus on first-order logic because of its simplicity and wide applicability to many reasoning problems, such as automated theorem proving~\citep{fitting2012first,gallier2015logic} and knowledge-based systems~\citep{van2008handbook}.
A separate planner module activates only a small subset of the meta-rules necessary for a given task and composes them into a deduction process (Section \ref{sec:PRIMA:planner}).
Thus, our planner-reasoner architecture features the dual capabilities of \emph{composing} and g{pruning} a logic deduction process, achieving a graceful capability-efficiency tradeoff in MTR. The model architecture is trained in an end-to-end manner using deep reinforcement learning (Section \ref{sec:learning}), and experimental results on several benchmarks demonstrate that this framework leads to a state-of-the-art balance between capability and efficiency (Section \ref{sec:exp}). We discuss related works in Section \ref{sec:related}, and conclude our paper in Section \ref{sec:conclusion}.

\section{Problem Formulation}
\label{sec:PRIMA:problem}

\begin{figure*}[htb!]
    \centering
    \includegraphics[width=1\textwidth]{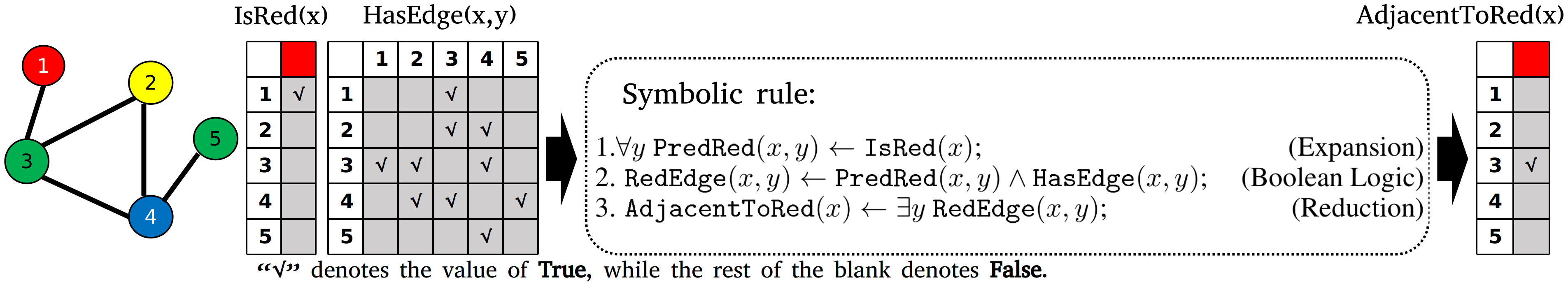}
    \caption{\footnotesize  An example from the $\mathtt{AdjacentToRed}$ task and its formulation as a logical reasoning problem.}
    \label{fig:atrexample}
\end{figure*}   

\paragraph{Logic reasoning} We begin with a brief introduction of the logic reasoning problem. Specifically, we consider a special variant of the First-Order Logic (FOL) system, which only consists of \emph{individual variables}, \emph{constants} and up to $r$-ary \emph{predicate variables}. That is, we do not consider \emph{functions} that map individual variables/constants into \emph{terms}. An $r$-ary predicate $p(x_1,\ldots,x_r)$ can be considered as a relation between $r$ constants, which takes the value of \texttt{True} or \texttt{False}. An atom $p(x_1, \cdots, x_r)$ is an $r$-ary predicate with its arguments $x_1, \cdots, x_r$ being either variables or constants. A \emph{well-defined formula} in our FOL system is a logical expression that is composed from atoms, logical connectives (e.g., negation $\neg$, conjunction $\wedge$, disjunction $\vee$,  implication $\leftarrow$), and possibly existential $\exists$ and universal $\forall$ quantifiers according to certain formation rules (see \citet{andrews2002introduction} for the details). In particular, the quantifiers $\exists$ and $\forall$ are only allowed to be applied to individual variables in FOL. In Fig.~\ref{fig:atrexample}, we give an example from the \texttt{AdjacentToRed} task~\citep{graves2016hybrid} and show how it could be formulated as a logical reasoning problem. Specifically, we are given a random graph along with the \emph{properties} (i.e., the color) of the nodes and the \emph{relations} (i.e., connectivity) between nodes. In our context, each node $i$ in the graph is a \emph{constant} and an \emph{individual variable} $x$ takes values in the set of constants $\{1,\ldots,5\}$. The properties of nodes and the relations between nodes are modeled as the unary predicate $\mathtt{IsRed}(x)$ ($5\times 1$ vector) and the binary predicate $\mathtt{HasEdge}(x,y)$ ($5\times 5$ matrix), respectively. The objective of logical reasoning is to deduce the value of the unary predicate $\mathtt{AdjacentToRed}(x)$ (i.e., whether a node $x$ has a neighbor of red color) from the \emph{base} predicates $\mathtt{IsRed}(x)$ and $\mathtt{HasEdge}(x,y)$ 
(see Fig.~\ref{fig:atrexample} for an example of the deduction process).

\paragraph{Multi-task reasoning}
Next, we introduce the definition of MTR.
With a slight abuse of notations, let $\{p(x_1,\ldots, x_r): r  \in [1,n]\}$ be the set of input predicates sampled from any of the $k$ different reasoning tasks, where $x_1, \ldots, x_r$ are the individual variables and $n$ is the maximum arity. A multi-task reasoning model takes $p(x_1,\ldots, x_r)$ as its input and seeks to predict the corresponding ground-truth output predicates $q(x_1,\ldots, x_r)$.
The aim is to learn multiple reasoning tasks jointly in a single model so that the reasoning skills in a task can be leveraged by other tasks to improve the general performance of all tasks at hand.

\section{PRIMA: Planner-Reasoner for Multi-task Reasoning}
\label{sec:PRIMA}
In this section, we develop our PRIMA framework, which is a novel neural-logic model architecture for multi-task reasoning. We begin with an introduction to the overall architecture (Section \ref{sec:PRIMA:multitask}) along with its key design insights, and then discuss its two modules in Sections \ref{sec:PRIMA:reasoner}--\ref{sec:PRIMA:planner}.

\begin{figure*}
    \centering
    \includegraphics[width=\textwidth]{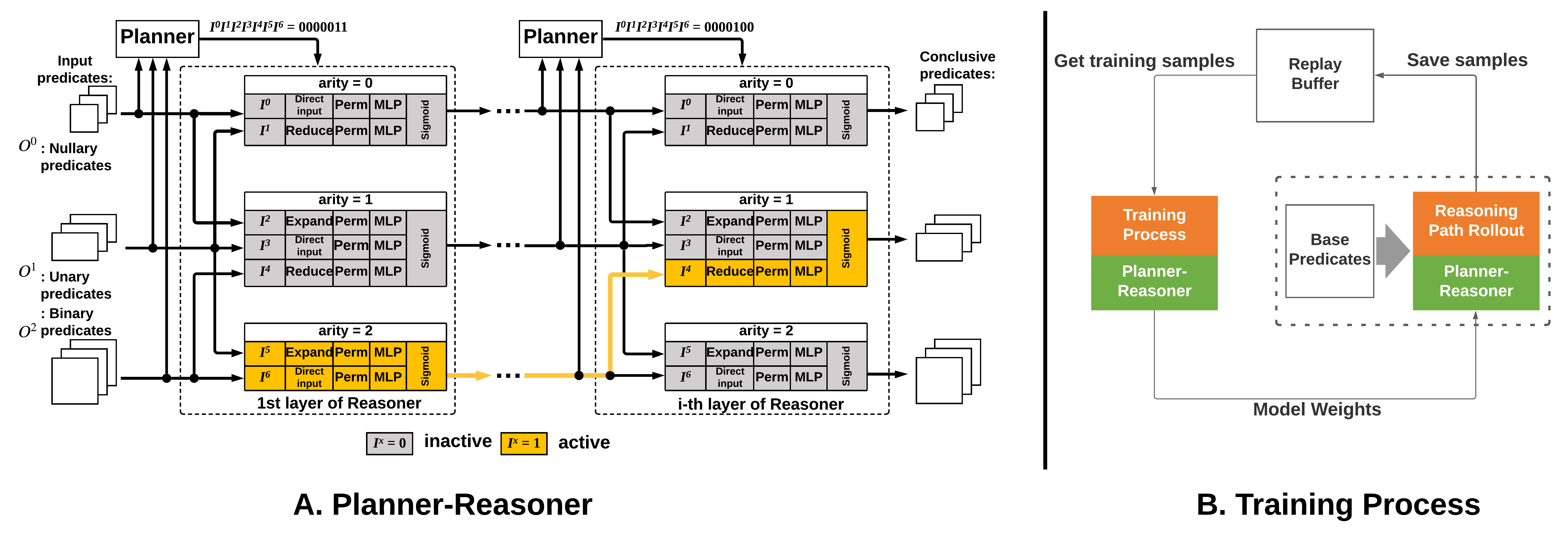}
    \caption{{\footnotesize Left (A): The Planner-Reasoner Architecture with an example solution. Right (B): The overall (end-to-end) training process of the model by deep reinforcement learning. ``Perm'' denotes $\mathrm{\texttt{Permute}}$ operator.}}
    \label{fig:reason-arch}
\end{figure*}

\subsection{The Overall Model Architecture of PRIMA}
\label{sec:PRIMA:multitask}
Logic reasoning typically seeks to chain an appropriate sequence of logic operators to reach desirable logical consequences from premises. There are two types of chaining to conduct logic reasoning: (i) forward-chaining and (ii) backward-chaining. The forward-chaining strategy recursively deduces all the possible conclusions based on all the available facts and deduction rules until it reaches the answer. In contrast, backward-chaining starts from the desired conclusion (i.e., the goal) and then works backward recursively to validate conditions through available facts. This paper adopts the forward-chaining strategy since the goal is unavailable here beforehand. Note that, in the MTR setting, it generally requires a large set of ``skills'' (logic operators) to tackle diverse tasks. Therefore, it is unrealistic to recursively generate all possible conclusions by chaining all the available deduction rules as in the vanilla forward-chaining strategy. Therefore, pruning the forward-chaining process is essential to improve MTR's reasoning efficiency.

We note that reasoning via pruned forward-chaining could be implemented by: (i) constructing the elementary logic operators and (ii) selecting and chaining a subset of these logic operators together. Accordingly, we develop a novel neural model architecture, named Planner-Reasoner (see Figure \ref{fig:reason-arch}A), to jointly accomplish these two missions with two interacting neural modules. Specifically, a ``Reasoner'' defines a set of neural-logic operators (representing learnable Horn clauses \citep{horn1951sentences}). Meanwhile, a ``Planner'' learns to select and chain a small subset of these logic operators together towards the correct solution (e.g., the orange-colored path in Figure \ref{fig:reason-arch}A). The inputs to the Planner-Reasoner model are the base predicates $\{p, p(x), p(x,y), \ldots\}$ that describe the premises. In particular, in the MTR setting, the nullary predicate $p$ characterizes the current task to be solved. This design allows the Planner to condition its decisions on the input task type. We feed the outputs (i.e., the conclusion predicates) into multiple output heads, one for each task (just as in standard multitask learning), to generate predictions. The entire model is trained in a complete data-driven end-to-end approach via deep reinforcement learning (Figure \ref{fig:reason-arch}B and Section \ref{sec:learning}): it trains the Planner to dynamically prune the forward-chaining process while jointly learning the logic operators in the Reasoner, leading to a state-of-the-art balance between MTR capability and efficiency (Section \ref{sec:exp}).

\subsection{Reasoner: transforming logic rules into neural operators}
\label{sec:PRIMA:reasoner}

The Reasoner module conducts logical deduction using a set of neural operators constructed from first-order logic rules (more specifically, a set of  ``learnable'' Horn clauses). Its architecture is inspired by NLM~\citep{nlm} (details about the difference can be found in Appendix \ref{sec:appendix:related:nlm}).
Three logic rules are considered as essential meta-rules: $\mathtt{Boolean Logic}$, $\mathtt{Expansion}$, and $\mathtt{Reduction}$.
$\mathtt{BooleanLogic:} ~\mathtt{expression}(x_1, x_2, \cdots, x_r) \rightarrow \hat{p}(x_1, x_2, \cdots, x_r)$,
where $\mathtt{expression}$ is composed of a combination of Boolean operations (\texttt{AND}, \texttt{OR}, and \texttt{NOT}) and $\hat{p}$ is the $\textit{output predicate}$.
For a given $r$-ary predicate and a given permutation $\psi \in S_n$, we define $p_\psi(x_1, \cdots, x_r) = p(x_{\psi(1)}, \cdots, x_{\psi(r)})$ where $S_n$ is the set of all possible permutations as the arguments to an input predicate. The corresponding neural implementation of $\mathtt{Boolean Logic}$ is $\sigma\left(\mathrm{{MLP}}\left(p_\psi(x_1, \cdots, x_r) \right);\theta \right)$, where $\sigma$ is the sigmoid activation function, $\mathrm{MLP}$ refers to a multi-layer perceptron, a $\mathrm{{Permute}}(\cdot)$ neural operator transforms input predicates to $p_\psi(x_1, \cdots, x_r)$, and $\theta$ is the learnable parameter within the model.
This is similar to the implicit Horn clause with the universal quantifier($\forall$), e.g., $p_1(x) \land p_2(x) \rightarrow \hat{p}(x)$ implicitly denoting $\forall x~ p_1(x) \land p_2(x) \rightarrow \hat{p}(x)$. 
The class of neural operators can be viewed as ``learnable'' Horn clauses.

$\mathtt{Expansion}$, and $\mathtt{Reduction}$ are two types of meta-rules for quantification that bridge predicates of different arities with logic quantifiers ($\forall$ and $\exists$).
$\mathtt{Expansion}$ introduces a new and distinct variable $x_{r+1}$ for a set of $r$-ary predicates with the universal quantifier($\forall$). For this reason, $\mathtt{Expansion}$ creates a new predicate $q$ from $p$.
$\mathtt{Expansion:}~ p(x_1, x_2, \cdots, x_{r})\rightarrow \forall{x}_{r+1}, q(x_1, x_2, \cdots, x_{r}, x_{r+1})$,
where  $x_{r+1} \notin \{x_i\}_{i=1}^{r}$.
The corresponding neural implementation of $\mathtt{Expansion}$, denoted by $\mathrm{Expand}(\cdot)$, expands the $r$-ary predicates into the $(r+1)$-ary predicates by repeating the $r$-ary predicates and stacking them in a new dimension.
Conversely, $\mathtt{Reduction}$ removes $x_{r+1}$ in a set of $(r+1)$-ary predicates via the quantifiers of $\forall$ or $\exists$.
$\mathtt{Reduction:} ~\forall {x_{r+1}}~p(x_1, x_2, \cdots, x_{r}, x_{r+1})
\rightarrow 
q(x_1, x_2, \cdots, x_{r}) 
, {\rm{or}~} \exists {x_{r+1}}~p(x_1, x_2, \cdots, x_{r}, x_{r+1})
\rightarrow 
q(x_1, x_2, \cdots, x_{r}) $.
The corresponding neural implementation of $\mathtt{Reduction}$, denoted by $\mathrm{Reduce}(\cdot)$, reduces the $(r+1)$-ary predicates into the $r$-ary predicates by taking the minimum (resp. maximum) along the dimension of $x_{r+1}$ for the universal quantifier $\forall$ (resp. existential quantifier $\exists$). 

\subsection{Planner: activating and forward-chaining learnable Horn clauses}
\label{sec:PRIMA:planner}
The Planner is our key module to address the capability-efficiency tradeoff in the MTR problem; it is responsible for activating the neural operators in the Reasoner and chaining them into reasoning paths. 
Existing learning-to-reason approaches, which are often based on inductive logic programming (ILP)~\citep{ilp:cropper2020turning,ilp30:cropper2020inductive,muggleton1994inductive} and the correspondingly neural-ILP methods~\citep{nlm,nlm:shi2020neural}. Conventional ILP methods suffer from several drawbacks, such as heavy reliance on human-crafted templates and sensitivity to noise. 
On the other hand, neural-ILP methods~\citep{nlm,nlm:shi2020neural}, leveraging the strength of deep learning and ILP,  such as the Neural Logic Machine (NLM)~\citep{nlm}, lack explicitness in the learning process of searching for reasoning paths. 
Let us take the learning process of NLM for example, which follows an intuitive two-step procedure. It first fully connects all the neural blocks and then searches all possible connections (corresponding to all possible predicate candidates) exhaustively to identify the desired reasoning path (corresponding to the desired predicate).

By using our proposed Planner module, we can strike a better capability-efficiency tradeoff.
Rather than conducting an exhaustive search over all possible predicate candidates as in NLM, the Planner prunes all the unnecessary operators and identifies an essential reasoning path with low complexity for a given problem.  Consider the following example. As shown in Fig.~\ref{fig:atrexample} and Fig.~\ref{fig:reason-arch}A (the highlighted orange-colored blocks), at each reasoning step, the Planner takes the input predicates and determines which neural operators should be activated. The decision is represented as a binary vector --- $[I^0 \ldots I^6]$ (termed \emph{operator footprint}) --- that corresponds to the neural operators of $\mathrm{Expand}$, $\mathrm{Reduce}$ and $\mathrm{DirectInput}$ at different arity groups. $\mathrm{DirectInput}$ is an identity mapping where its following operations of $\mathrm{Permute}$, $\mathrm{MLP}$ and sigmoid nonlinearity can directly approximate the $\mathtt{Boolean Logic}$.
By chaining these sequential decisions, a sequence of operator footprints is formulated as a reasoning path, as the highlighted orange-colored path in Fig.~\ref{fig:reason-arch}A.
Generally, the neural operators defined in the Reasoner can also be viewed as (learnable) Horn Clauses~\citep{horn1951sentences,chandra1985horn}, and the Planner \emph{forward-chains} them into a reasoning path.

\section{Learning-to-Reason via Reinforcement Learning}
\label{sec:learning}
In our problem, the decision variables of the Planner are binary indicators of whether a neural operator module should be activated. 
Readers familiar with Markov decision processes (MDPs)~\citep{puterman} might notice that the reasoning path of our model in Fig.~\ref{fig:reason-arch}A resembles a temporal rollout in MDP formulation~\citep{sutton2018reinforcement}.
Therefore, we frame learning-to-reason as a sequential decision-making problem and adopt off-the-shelf reinforcement learning (RL) algorithms.

\subsection{An MDP formulation of the learning-to-reason problem}
\label{sec:rl:formulation}
An MDP is defined as the tuple $({\mathcal{S},\mathcal{A},P_{ss'}^{a},R,\gamma})$, where $\mathcal{S}$ and $\mathcal{A}$ are finite sets of states and actions, the transition kernel $P_{ss'}^{a}$ specifies the probability of transition from state $s\in\mathcal{S}$ to state $s'\in\mathcal{S}$ by taking action $a\in\mathcal{A}$, $R(s,a):\mathcal{S}\times\mathcal{A}\to\mathbb{R}$ is the reward function, and $0\leq\gamma\leq1$ is a discount factor. A stationary policy $\pi:\mathcal{S}\times\mathcal{A}\to\left[{0,1}\right]$ is a probabilistic mapping from states to actions. The primary objective of an RL algorithm is to identify a near-optimal policy $\pi^*$ that satisfies 
$\pi^* := \argmax_\pi \bigg\{ 
J(\pi) := \mathbb{E}\bigg[\sum\limits_{t = 0}^{T_{\max} - 1} {\gamma^t r({s_t},{a_t\sim \pi})}\bigg]
\bigg\}$,
where $T_{\max}$ is a positive integer denoting the horizon length --- that is, the maximum length of a rollout.
The resemblance between Fig.~\ref{fig:reason-arch} and an MDP formulation is relatively intuitive as summarized in Table~\ref{tab:rl-planner}. At the $t$-th time step, the state $s_t$ corresponds to the set of predicates, $s_t = [O^{0}_{t}, O^{1}_{t}, O^{2}_{t}]$, with the superscript denoting the corresponding arity. The action $a_t = [I_{t}^0, I_{t}^1, \ldots, I_{t}^{K-1}]$ (e.g., $a_0 = [000011]$) is a binary vector that indicates the activated neural operators (i.e., the operator footprint), where $K$ is the number of operators per layer.
The reward $r_t$ is defined to be
\begin{align}
    r_t := 
\begin{cases}
    -\sum_{i=0}^{K-1} I_{t}^i,& \text{if } t< T_{\max}\\
    \mathtt{Accuracy},              & t = T_{\max}
\end{cases}.
\label{eq:r}
\end{align}
That is, the terminal reward is set to be the reasoning accuracy at the end of the reasoning path (see Appendix \ref{appendix:algorithm_details:acc} for its definition), and the intermediate reward at each step is chosen to be the negated number of activated operators (which penalizes the cost of performing the current reasoning step).
The transition kernel $P_{ss'}^{a}$ corresponds to the function modeled by one Reasoner layer; each Reasoner layer will take the state (predicates) $s_{t-1}=[O_{t-1}^0,O_{t-1}^1,O_{t-1}^2,]$ and the action (operator footprint) $a_t=[I_{t}^0, I_{t}^1, \ldots, I_{t}^{K-1}]$ as its input and then generate the next state (predicates) $s_{t}=[O_{t}^0,O_{t}^1,O_{t}^2,]$. This also implies that the Reasoner layer defines a deterministic transition kernel, i.e., given $s_{t-1}$ and $a_t$ the next state $s_t$ is determined.

\begin{table*}[htb!]
\caption{The identification between the concepts of PRIMA and that of RL at the $t$-th time step.}
\label{tab:rl-planner}
\centering
{\small
\begin{tabular}{c|c|c|c|c|c|c}
\toprule
{\small RL} &
  {\small State} $s_t$ &
  {\small Action} $a_t$ &
  {\small Reward} $r_t$ &
  \begin{tabular}[c]{@{}c@{}}{\small Transition}\\ {\small Kernel} $P_{ss'}^{a}$\end{tabular} &
  {\small Policy} &
  {\small Rollout}  
 \\ \midrule
\begin{tabular}[c]{@{}c@{}} {\small PRIMA} \end{tabular} &
  \begin{tabular}[c]{@{}c@{}}{\small Predicates of different arities:} \\ $[O^0_t, O^1_t, O^2_t]_t$\end{tabular} &
  \begin{tabular}[c]{@{}c@{}}{\small Operator footprint:}\\ $[I_t^0 \ldots I_t^{K-1}]$\end{tabular} &
  Eq.~(\ref{eq:r}) &
  \begin{tabular}[c]{@{}c@{}}{\small One layer of}\\ {\small \textbf{Reasoner}}\end{tabular} &
  \begin{tabular}[c]{@{}c@{}} {\small \textbf{Planner}} \end{tabular} &
  {\small Reasoning path}
  \\ \bottomrule
\end{tabular}%
}
\end{table*}

\begin{figure*}[htb!]
    \centering
    \includegraphics[width=.86\textwidth]{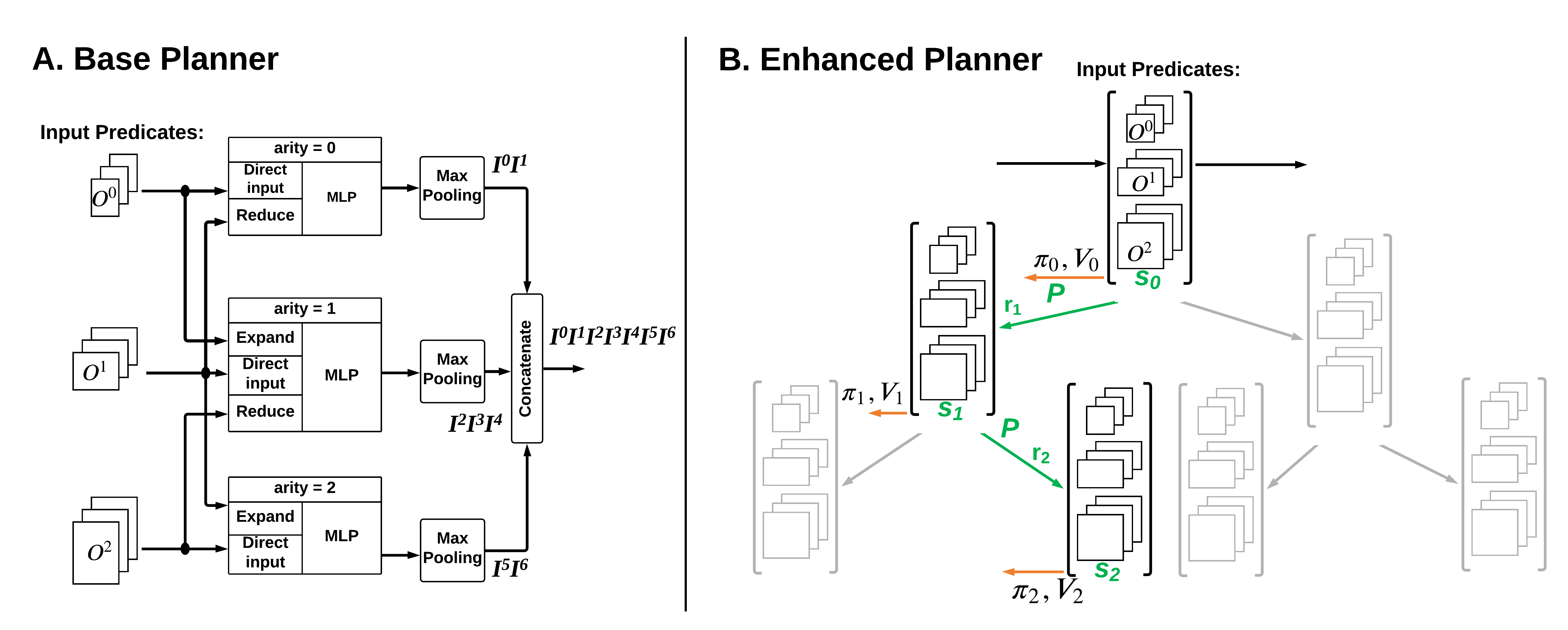}
    \caption{{\footnotesize The architecture of the Base Planner and its enhanced version. Left: The Base Planner is based on a one-layer fully-activated Reasoner followed by max-pooling operations to predict an indicator of whether an op should be activated. Right: The Enhanced Planner uses MCTS to further boost the performance.}}
    \label{fig:plan-arch}
\end{figure*}  

\subsection{Policy network: modeling of planner}
\label{sec:planner-model}
The Planner module is embodied in the policy network. As shown in Fig.~\ref{fig:plan-arch}A, the base planner is a separate module that has the same architecture of 1-layer (fully-activated) Reasoner followed by a max-pooling layer.
This architecture enables the reduction of input predicates to the specific indicators by reflecting whether the operations at the corresponding position of one layer of Reasoner are active or inactive.
Further, we can also leverage Monte-Carlo Tree Search (MCTS)~\citep{mcts:browne2012survey,mcts:munos2014bandits} to boost the performance, which leads to an Enhanced Planner (Fig.~\ref{fig:plan-arch}B). An MCTS algorithm such as the Upper Confidence Bound for Trees (UCT) method ~\citep{kocsis2006improved}, is a model-based RL algorithm that plans the best action at each time step \citep{mcts:browne2012survey} by constructing a search tree, with states as nodes and actions as edges. The Enhanced Planner uses MCTS to exploit partial knowledge of the problem structure (i.e., the deterministic transition kernel $P_{ss'}^a$ defined by the Reasoner layer) and construct a search tree to help identify the best actions (which ops to activate).
Details of the MCTS algorithms used in the Enhanced Planner can be found in Appendix~\ref{appendix:algorithm_details:mcts}.

\subsection{Overall learning framework}
\label{sec:planner:overall}
As illustrated in Fig.~\ref{fig:plan-arch}, we introduce concrete data-driven decision-making methods---that is, RL approaches~\citep{sutton2018reinforcement}---to address the learning-to-reason problem. To illustrate this, we apply the model-free RL method REINFORCE~\citep{williams1992simple} and the model-based method MuZero~\citep{muzero}. 
Compared with model-free reinforcement learning, model-based reinforcement learning (MBRL) more effectively handles large search-space problems, such as the game of Go~\citep{alphagozero,alphagozero,alphazero}. MuZero~\citep{muzero}, a recently proposed MBRL approach to integrating planning and learning, has achieved great success with a variety of complex tasks. Motivated by the success of MuZero, we propose an MBRL approach for neural-symbolic reasoning. 
The key insight behind adopting MuZero is that in real applications, we typically have partial structural knowledge of the transition kernel $P_{ss'}^{a}$ and reward function $r(s,a)$. 
As a result of the model-based module, testing complexity can be greatly reduced by adopting MCTS, which leads to a better set of predicates.
Of course, MuZero is just one option in the model-based family of approaches. We leave it as future research to propose and compare other model-based alternatives.

The pipeline of the training process is illustrated in Fig.\ref{fig:reason-arch}.B (the right subfigure). After loading the model weights, the reasoning path rollouts are executed by the agent (or model instance), according to the current policy network. The performed reasoning path rollout is then stored in the replay buffer. The Planner-Reasoner is trained via rollouts sampled from the replay buffer.

\section{Experimental Results and Analysis}
\label{sec:exp}

In this section, we evaluate the performance of different variants of PRIMA  on eight tasks from the family tree and graph benchmarks~\citep{graves2016hybrid}, including \texttt{1-Outdegree}, \texttt{AdjacentToRed}, \texttt{HasFather}, \texttt{HasSister}, \texttt{4-Connectivity}, \texttt{IsGrandparent}, \texttt{IsUncle}, \texttt{IsMGUncle}. 
These tasks are widely used benchmarks for inductive logic programming~\citep{krotzsch2020computing,calautti2015chase}. Detailed descriptions about those tasks can be found in Appendix~ \ref{appendix:experiment_detail:benchmark}.
We evaluate their testing accuracy and reasoning cost (measured in FLOPs: the number of floating-point operations executed \citep{clark2020electra}) on these tasks and compare them to several baselines. Furthermore, detailed case studies are conducted on the reasoning path, which indicates the operator sharing among different tasks. All the results demonstrate the graceful capability-efficiency tradeoff of PRIMA in MTR.

\subsection{Experimental Setups}
In the multi-task setting, we first randomly sample a task according to a pre-defined probability distribution (over different tasks), and then generate the data for the selected task using the same methods as in NLM~\citep{nlm}. In addition, we also augment the generated data with one-hot encoding to indicate the selected task, which will be further converted into nullary (background) input predicates. Also, task-specific output heads are introduced to generate outputs for different tasks. These adaptions apply to DLM-MTR and PRIMA.
The reasoning accuracy is used as the reward. In the inference (or testing) stage, the Reasoner is combined with the learned Base Planner to perform the tasks, instead of an enhanced planner (MCTS), to reduce the extra computation.
The problem size for training is always $10$ for graph tasks and $20$ for family tree tasks across all settings, regardless of the sizes of testing problems.

\subsection{Overall performance}
\paragraph{Single-task reasoning capability}
Before we proceed to evaluate the MTR capabilities of PRIMA, we first adapt it to the single-task setting by letting the nullary input predicates to be zero, and then train a separate model for each task. We name this single-task version of PRIMA as PRISA (i.e., \textit{\textbf{P}lanner-\textbf{R}easoner \textbf{I}nside a \textbf{S}ingle Reasoning \textbf{A}gent}), and compare it with existing (single-task) neural-logic reasoning approaches (e.g., NLM~\citep{nlm}, DLM~\citep{matthieu2021differentiable}, and MemNN~\citep{sukhbaatar2015end}).
First, we compare the performance of three different variants of PRISA (Section~\ref{sec:planner:overall}) for single-task reasoning, which learns their planners based on different reinforcement learning algorithms; PRISA-REINFORCE uses REINFORCE~\citep{williams1992simple}, PRISA-PPO uses PPO~\citep{schulman2017proximal}, and PRISA-MuZero uses MuZero~\citep{muzero}.
We report the test accuracy and the Percentage of Successful Seeds (PSS) in Table~\ref{tab:x5} to measure the model's reasoning \emph{capabilities}, where the PSS reaches 100\% of success rates \citep{matthieu2021differentiable}.
We note that PRISA-MuZero has the same 100\% accuracy as NLM~\citep{nlm}, DLM~\citep{matthieu2021differentiable}, and $\partial$ILP~\citep{evans2018learning} across different tasks, and outperforms MemNN~\citep{sukhbaatar2015end} (shown in the single-task part in Table~\ref{tab:x6}). 
But it also has a higher successful percentage (PSS) compared to other methods. The results show that PRISA-MuZero achieves the best performance on single-task reasoning, which confirms the strength of the MCTS-based Enhanced Planner (Section~\ref{sec:planner:overall}) and the MuZero learning strategy. Therefore, we will use the Enhanced Planner and MuZero in PRISA and PRIMA in the rest of our empirical studies.

\paragraph{Multi-task reasoning capability}
Next, we evaluate the MTR capabilities of PRIMA. To the best of our knowledge, there is no existing approach designed specifically for MTR.  
Therefore, we adapt DLM into its multi-task versions, named  DLM-MTR, respectively. DLM-MTR  follow the same input and output modification as what we did to upgrade PRISA to PRIMA (Section~\ref{sec:PRIMA:multitask}).
By this, we can examine the contribution of our proposed \emph{Planner-Reasoner architecture} for MTR. As shown in Table~\ref{tab:x6}, PRIMA (with MuZero as the Base Planner) performs well (perfectly) on different reasoning tasks. 
On the other hand, DLM-MTR experiences some performance degradation (on \texttt{AdjacentToRed}). This result confirms that our Planner-Reasoner architecture is more suitable for MTR. We conjecture that the benefit comes from using a Planner to explicitly select the necessary neural operators for each task, avoiding potential conflicts between different tasks during the learning process.

\begin{figure*}[htb!]
    \centering
    \includegraphics[width=.95\textwidth,height=3.2cm]{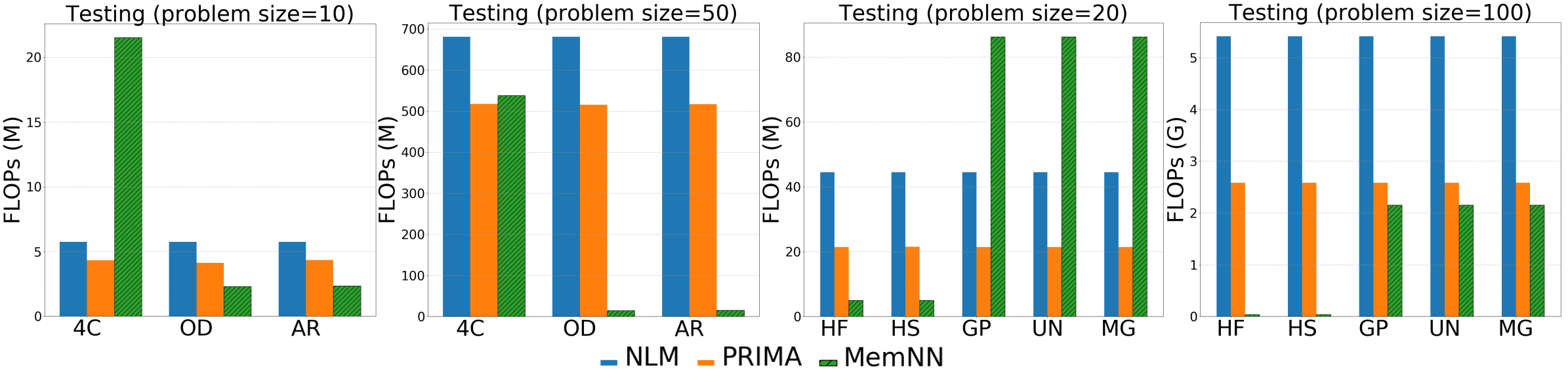}
    \caption{{\footnotesize The reasoning costs (in FLOPs) of different models at the inference stage (1 M=1$\times 10^6$, 1 G=1$\times 10^9$). Compared to NLM family, our PRIMA significantly reduces the reasoning complexity by intelligently selecting ops (short for neural operators) using a planner. Although the FLOPs of MemNN seem low in most cases of testing, its testing accuracy is bad and cannot achieve accurate predictions (see Table~\ref{tab:x6}).
    ``OD'' denotes \texttt{1-Outdegree}, likewise,
    ``AR'':\texttt{AdjacentToRed}, ``4C'':\texttt{4-Connectivity}, ``HF'':\texttt{HasFather}, ``HS'':\texttt{HasSister}, ``GP'':\texttt{IsGrandparent}, ``UN'':\texttt{IsUncle}, ``MG'':\texttt{IsMGUncle}.
    }
}
    \label{fig:comp-flops1}
\end{figure*}

Experiments are also conducted to test the performance of PRIMA with different problem sizes.
The problem size in training is 10 for all graph tasks and 20 for all family-tree tasks. 
In testing, we evaluate the methods on much larger problem sizes (50 for the graph tasks and 100 for the family tree tasks), which the methods have not seen before. Therefore, the Planner must dynamically activate a proper set of neural operators to construct a path to solve the new problem. As reported in Fig.~\ref{fig:comp-flops1} and Tables~\ref{tab:x5} and \ref{tab:x6}, PRIMA can achieve the best accuracy and lower flops when the problem sizes for training and testing are different.

\begin{table*}[]
\caption{{\footnotesize \textbf{Testing Accuracy and PSS} of different variants of PRISA on different tasks. PRISA-MuZero achieves the best performance on single-task reasoning, which confirms the strength of the MCTS-based Enhanced Planner and the MuZero learning strategy. ``\textit{m}'': the problem size. ``PSS'': Percentage of Successful Seeds.}}
\label{tab:x5}
\resizebox{\textwidth}{!}{%
\begin{tabular}{|c|c|c|c|ccccc|c|ccc|}
\hline
\multirow{13}{*}{{\rotatebox[origin=c]{90}{testing accuracy}}} & \multicolumn{2}{c|}{} & \begin{tabular}[c]{@{}c@{}}Family\\ Tree\end{tabular} & HasFather & HasSister & IsGrandparent & IsUncle & IsMGUncle & Graph & AdjacentToRed & 4-Connectivity & 1-OutDegree \\ \cline{2-13} 
 & \multirow{9}{*}{{\rotatebox[origin=c]{90}{Single Task}}} & \multirow{3}{*}{\begin{tabular}[c]{@{}c@{}}PRISA-\\ REINFORCE\end{tabular}} & m=20 &62.6 &50.7 &96.5 &97.3 &99.8 & m=10 &47.7 &33.5 &48.7 \\
 & & & m=100 &87.8 &69.8 &2.3 &97.7 &98.4 & m=50 &71.6 &92.8 &97.4 \\
 & & & PSS &0 &0 &0 &0 &0 & PSS &0 &0 &0 \\ \cline{3-13} 
 & & \multirow{3}{*}{PRISA-PPO} & m=20 &71.5 &64.3 &97.5 &98.1 &99.6 & m=10 &62.3 &57.8 &61.6 \\
 & & & m=100 &93.2 &78.7 &98.2 &97.3 &99.1 & m=50 &85.5 &95.2 &96.3 \\
 & & & PSS &0 &0 &0 &0 &0 & PSS &0 &0 &0 \\ \cline{3-13} 
 & & \multirow{3}{*}{PRISA-MuZero} & m=20 &\bf{100} &\bf{100} &\bf{100} &\bf{100} &\bf{100} & m=10 &\bf{100} &\bf{100} &\bf{100} \\
 & & & m=100 &\bf{100} &\bf{100} &\bf{100} &\bf{100} &\bf{100} & m=50 &\bf{100} &\bf{100} &\bf{100} \\
 & & & PSS &\bf{100} &\bf{100} &\bf{100} &\bf{100} &\bf{100} & PSS &90 &\bf{100} &\bf{100} 

\\ \hline
\end{tabular}%
}
\end{table*}

\begin{table*}[htb!]
\caption{\small{\textbf{Testing Accuracy and PSS} of PRIMA and other baselines on different reasoning tasks. The results of $\partial$ILP, NLM, and DLM are merged in one row due to space constraints and are presented in the same order if the results are different. 
``\textit{m}'': the problem size. ``PSS'': Percentage of Successful Seeds. Numbers in red denote $< 100\%$.}}
\resizebox{\textwidth}{!}{%
\begin{tabular}{|c|c|c|c|ccccc|c|ccc|}
\hline
\multirow{16}{*}{{\rotatebox[origin=c]{90}{testing accuracy}}} & \multicolumn{2}{c|}{} & \begin{tabular}[c]{@{}c@{}}Family \\Tree \end{tabular} & HasFather & HasSister & IsGrandparent & IsUncle & IsMGUncle & Graph & AdjacentToRed & 4-Connectivity & 1-OutDegree \\ \cline{2-13} 
 & \multirow{6}{*}{{\rotatebox[origin=c]{90}{Single Task}}} & \multirow{3}{*}{MemNN} & m=20 &{\color{red}99.9} &{\color{red}86.3} &{\color{red}96.5} &{\color{red}96.3} &{\color{red}99.7} & m=10 &{\color{red}95.2} &{\color{red}92.3} &{\color{red}99.8} \\
 & & & m=100 &{\color{red}59.8} &{\color{red}59.8} &{\color{red}97.7} &{\color{red}96} &{\color{red}98.4} & m=50 &{\color{red}93.1} &{\color{red}81.3} &{\color{red}78.6} \\
 & & & PSS &{\color{red}\textbf{0}} &{\color{red}\textbf{0}} &{\color{red}\textbf{0}} &{\color{red}\textbf{0}} &{\color{red}\textbf{0}} & PSS &{\color{red}\textbf{0}} &{\color{red}\textbf{0}} &{\color{red}\textbf{0}} \\ \cline{3-13} 
 & & \multirow{3}{*}{$\partial$ILP/ NLM/ DLM} & m=20 &100 &100 &100 &100 &100 & m=10 &100 &100 &100 \\
 & & & m=100 &100 &100 &100 &100 &100 & m=50 &100 &100 &100 \\
 & & & PSS &100 &100 &100 &{100/ {\color{red}90}/ 100} &{100/ {\color{red}20}/ {\color{red}70}} & PSS &{100/ {\color{red}90}/ {\color{red}90}} &100 &100 \\
 \cline{2-13}
 & \multirow{6}{*}{{\rotatebox[origin=c]{90}{Multi-Task}} {\rotatebox[origin=c]{90}{}}} & 
 \multirow{3}{*}{DLM-MTR} & m=20 &100 &100 &100 &100 &100 & m=10 &{\color{red}96.7} &100 &100 \\
 & & & m=100 &100 &100 &100 &100 &100 & m=50 &{\color{red}97.2} &100 &100 \\
 & & & PSS &100 &100 &100 &100 &100 & PSS &{\color{red}\textbf{0}} &100 &100 \\ \cline{3-13} 
 & & \multirow{3}{*}{PRIMA} & m=20 &100 &100 &100 &100 &100 & m=10 &100 &100 &100 \\
 & & & m=100 &100 &100 &100 &100 &100 & m=50 &100 &100 &100 \\
 & & & PSS &100 &100 &100 &100 &{\color{red}90} & PSS &{\color{red}90} &100 &100 \\ \hline
\end{tabular}%
}
\label{tab:x6}
\end{table*}

\begin{figure*}[htb!]
    \centering
    \includegraphics[width=.95\textwidth]{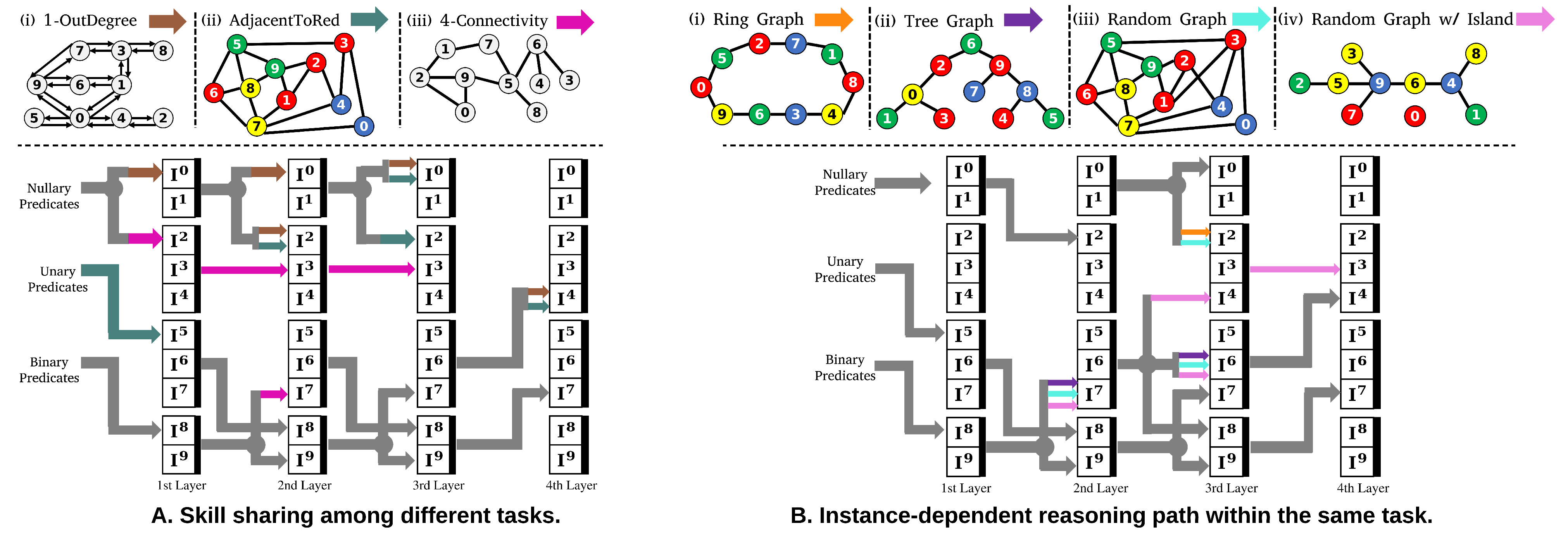}
    \caption{\footnotesize  The reasoning paths of PRIMA. The gray arrows denote the ``shared path''. The colored arrows in sub-figures (A) and (B) denote task-specific paths and instance-specific paths, respectively. In (A), besides the \texttt{AdjacentToRed} task we have introduced earlier, we also consider the \texttt{1-Outdegree} task, which reasons about whether the out-degree of a node is exactly equal to $1$, and the \texttt{4-Connectivity} task, which is to decide whether there are two nodes connected within $4$ hops.
    }
    \label{fig:reason-path}
\end{figure*}

\paragraph{Operator/Path sharing in MTR}
To take a closer look into how PRIMA achieves such a better capability-efficiency tradeoff (in Fig. \ref{fig:comp-flops1}), we examine the reasoning paths on three different graph tasks: \texttt{1-Outdegree}, \texttt{AdjacentToRed}, and \texttt{4-Connectivity}. Specifically, we sample instances from these three tasks and feed them into PRIMA separately to generate their corresponding reasoning paths. The results are plotted in Fig. \ref{fig:reason-path}A, where the gray paths denote the ones shared across tasks, and the colored ones are task-specific. It clearly shows that PRIMA learns a large set of neural operators sharable across tasks. Given each input instance from a particular task, PRIMA activates a set of shared paths along with a few task-specific paths to deduce the logical consequences.

\paragraph{Generalizability of the Planner}
To demonstrate the generalizability of the Planner module in PRIMA, we generate input instances of \texttt{AdjacentToRed} with topologies that have not been seen during training. In Fig.~\ref{fig:reason-path}B, we show the reasoning paths activated by the Planner for these different topologies, which demonstrates that different input instances share a large portion of reasoning paths. This fact is not surprising as solving the same task of \texttt{AdjacentToRed} should rely on a common set of skills. More interestingly, we notice that even for solving this same task, the Planner will have to call upon a few \emph{instance-dependent} sub-paths to handle the subtle inherent differences (e.g., the graph topology) that exist between different input instances.
For this reason, PRIMA maintains an instance-dependent \textit{dynamic} architecture, which is in sharp contrast to the Neural Architecture Search (NAS) approaches. Although NAS may also use RL algorithms to seek for a smaller architecture~\citep{zoph2017neural}, it only searches for a \emph{static} architecture that will be applied to all the input instances.


\section{Related Work}
\label{sec:related}

\paragraph{Multi-task learning}
Multi-task Learning~\citep{zhang2021survey,zhou2011malsar,pan2009survey} has focused primarily on the supervised learning paradigm, which itself can be divided into several approaches. The first is feature-based MTL, also called multi-task feature learning,~\citep{aep08}, which explores the sharing of features across different tasks using regularization techniques~\citep{shinohara16,lqh17}. The second approach assumes that tasks are intrinsically related, such as low-rank learning ~\citep{az05,zgy05}, learning with task clustering~\citep{glh11,zzl16}, and task-relation learning (such as via task similarity, task correlation, or task covariance)~\citep{gzb16,zhang13,cmpr15}.

\paragraph{Neural-symbolic reasoning} 
Neural-symbolic AI for reasoning and inference has a long history~\citep{nsai:besold2017neural,nsai:bader2004integration,nsai:garcez2008neural}, and neural-ILP has developed primarily in two major directions ~\citep{ilp:cropper2020turning}. 
The first direction applies neural operators such as tensor calculus to simulate logic reasoning~\citep{nsai:yang2017differentiable,nlm,nlm:shi2020neural} and~\citep{nlm:manhaeve2018deepproblog}.
The second direction involves relaxed subset selection~\citep{evans2018learning,nsrl:si2019} with a predefined set of task-specific logic clauses. This approach reduces the task to a subset-selection problem by selecting a subset of clauses from the predefined set and using neural networks to search for a relaxed solution.

Our work is different from the most recent work, such as~\citep{nlm,nlm:shi2020neural}, in several ways. 
Compared with~\citep{nlm}, the most notable difference is the improvement in efficiency by introducing learning-to-reason via reinforcement learning. A second major difference is that PRIMA offers more generalizability than NLM by decomposing the logic operators into more finely-grained units.
We refer readers for more detailed related work to Appendix~\ref{sec:appendix:related:nlm}.

\section{Conclusion}
\label{sec:conclusion}
A long-standing challenge in multi-task reasoning (MTR) is the intrinsic conflict between capability and efficiency. 
Our main contribution is the development of a novel neural-logic architecture termed PRIMA, which learns to perform efficient multitask reasoning (in first-order logic) by dynamically chaining learnable Horn clauses (represented by the neural logic operators). PRIMA improves inference efficiency by dynamically pruning unnecessary neural logic operators and achieves a state-of-the-art balance between MTR capability and inference efficiency. The model's training follows a complete data-driven end-to-end approach via deep reinforcement learning, and the performance is validated across a variety of benchmarks. Future work could include extending the framework to high-order logic and investigating scenarios when meta-rules have a hierarchical structure.

\bibliographystyle{apalike}
\bibliography{ref,mtl}

\clearpage
\begin{center}
{\Large \textbf{Appendix}}
\end{center}

\appendix
\input{appendix}

\end{document}

%% file: math_commands.tex
\usepackage{amsmath,amsfonts,bm}

\def\eqref#1{equation~\ref{#1}}

\def\1{\bm{1}}

\DeclareMathAlphabet{\mathsfit}{\encodingdefault}{\sfdefault}{m}{sl}
\SetMathAlphabet{\mathsfit}{bold}{\encodingdefault}{\sfdefault}{bx}{n}

\DeclareMathOperator*{\argmax}{arg\,max}

%% file: appendix.tex
\section{Algorithm Details}
\label{appendix:algorithm_details}

\subsection{Details of MCTS: four key steps}
\label{appendix:algorithm_details:mcts}
During each epoch, MCTS repeatedly performs four sequential steps: selection, expansion, simulation, and backpropagation. The selection step traverses the existing search tree until the leaf node (or termination condition) by choosing actions (edges) $a_s$ at each node $s$ according to a tree policy. 
One widely used tree policy is the UCT~\citep{kocsis2006improved} policy, which conducts the node-selection via
\begin{align}
    a_{s} = \argmax_{s' \in \mathcal{C}(s)} 
    \bigg\{ V_{s'} + \beta \sqrt{\frac{2 \log N_{s}}{N_{s'}}}  \bigg\},
    \label{Eq:uct_select_action}
\end{align}
where $\mathcal{C}(s)$ denotes the set of all child nodes for $s$, the first term $V_{s'}$ is an estimate of the long-term cumulative reward that can be received when starting from the state represented by node $s'$, and the second term represents the uncertainty (confidence-interval size) of that estimate. The confidence interval is calculated based on the upper confidence bound (UCB) \citep{auer2002finite,auer2002using} using $N_s$ and $N_{s'}$, which denote the number of times that nodes $s$ and $s'$ have been visited (respectively). 
The key idea of UCT policy (Eq.~(\ref{Eq:uct_select_action})) is to select the best action according to an optimistic estimation (the upper confidence bound) of the expected return, which balances the exploitation (first term) and exploration (second term) with $\beta$ controlling the trade-off.
The second step (node expansion) is conducted according to a prior policy by adding a new child node if the selection process reaches a leaf node of the search tree.
Next, the simulation step estimates the value function (cumulative reward) $\hat{V}_s$ by running the environment simulator with a default (simulation) policy. Finally, the \emph{backpropagation} step updates the statistics $V_{\state}$ and $N_{\state}$ from the leaf node $\state_{T}$ to the root node $\state_{0}$ of the selected path by recursively performing the following update (i.e., from $t = T - 1$ to $t = 0$):
    \begin{align}
        N_{\state_{t}} \leftarrow N_{\state_{t}} + 1, \quad 
        \hat{V}_{\state_{t}} \leftarrow r (\state_{t}, \action_{t}) + \gamma \hat{V}_{\state_{t + 1}}, \quad
        V_{\state_{t}} \leftarrow \big ( ( N_{\state_{t}} - 1 ) V_{\state_{t}} + \hat{V}_{\state_{t}} \big ) / N_{\state_{t}},
        \label{eq:backpropagation}
    \end{align}
\noindent where $\hat{V}_{\state_{T}}$ is the simulation return of $\state_{T}$, and $\action_{t}$ denotes the action selected following Eq.~(\ref{Eq:uct_select_action}) at state $\state_{t}$.
Generally, the expansion step and the simulation step
are more time-consuming than the other two steps, as they involve a large number of interactions with the environment.

In our Enhanced Planner, we adopt a recently proposed variant of MCTS that is based on probabilistic upper confidence tree (PUCT)~\citep{rosin2011multi}. It conducts node-selection according to a so-called {\texttt{PUCT}} score with two components ${\texttt{PUCT}}(s, a) = Q(s, a) + U(s, a)$. $Q(s,a)$ is the mean state-action value calculated from the averaged game result that takes action $a$ during current simulations, and $U(s,a)$ is the exploration bonus calculated as
\begin{align}
U(s,a) = \pi(a|s)\frac{{\sqrt {\sum\nolimits_b {N(s,b)} } }}{{1 + N(s,a)}}\bigg(c_1 + \log\Big(\frac{\sum_b N(s, b) + c_2 + 1}{c_2}\Big)\bigg),
\end{align}
where $a,b$ are possible actions, $\pi(a|s)$ is the output of the Base Planner, $N(s,a)$ is the visit count of the $(s,a)$ pair during current simulations, and constants $c_1, c_2$ are the exploration-controlling hyper-parameters.

An MCTS algorithm is a model-based RL algorithm that plans the best action at each time step \citep{mcts:browne2012survey} by constructing a search tree, with states as nodes and actions as edges. 
It uses the MDP model to identify the best action at each time step until the leaf node (or until other termination conditions are satisfied) by choosing actions (edges) $a_s$ at each node $s$ according to a tree policy. 
For implementations of Monte-Carlo Tree Search, we refer readers to \url{https://github.com/werner-duvaud/muzero-general} for details.

\subsection{Details of calculating reasoning accuracy}
\label{appendix:algorithm_details:acc}
Since the ``target'' predicates are available for all objects or pairs of objects, the reasoning accuracy refers to the accuracy evaluated on all objects (for properties such as \texttt{AdjacentToRed(x)} or \texttt{1-OutDegree(x)}) or pairs of objects (for relations such as \texttt{4-Connectivity(x, y)}).

\section{Experimental Details}
\label{appendix:experiment_details}

\subsection{Reasoning tasks introduction}
\label{appendix:experiment_detail:benchmark}
\paragraph{Graph} For graph tasks, they have the same background knowledge (or background predicate): \texttt{HasEdge}$(x, y)$, i.e., \texttt{HasEdge}$(x, y)$ is \textit{True} if there is an undirected edge between node $x$ and node $y$. However, \texttt{AdjacentToRed} has an extra background predicate, \texttt{IsRed}$(x)$. \texttt{IsRed}$(x)$ is \textit{True} if the color of node $x$ is red. Specifically, these graph tasks seeks to predict the target concepts (or target predicates) shown as below.
\begin{itemize}
    \item \texttt{1-OutDegree}: \texttt{1-OutDegree}$(x)$ is \textit{True} if the out-degree of node $x$ is exactly $1$.
    \item \texttt{AdjacentToRed}: \texttt{AdjacentToRed}$(x)$ is \textit{True} if the node $x$ has an edge with a red node.
    \item \texttt{4-Connectivity}: \texttt{4-Connectivity}$(x, y)$ is \textit{True} if there are within $4$ hops between node $x$ and node $y$.
\end{itemize}

\paragraph{Family Tree} For family tree tasks, they have the same background knowledge (or background predicates):  \texttt{IsFather}$(x, y)$, \texttt{IsMother}$(x, y)$, \texttt{IsSon}$(x, y)$ and \texttt{IsDaughter}$(x, y)$. For instance, \texttt{IsFather}$(x, y)$ is \textit{True} when $y$ is $x$’s father. Specifically, these family tree tasks seek to predict the target concepts (or target predicates) shown below.
\begin{itemize}
    \item \texttt{HasFather}: \texttt{HasFather}$(x)$ is \textit{True} if $x$ has father.
    \item \texttt{HasSister}: \texttt{HasSister}$(x)$ is \textit{True} if $x$ has at least one sister.
    \item \texttt{IsGrandparent}: \texttt{IsGrandparent}$(x, y)$ is \textit{True} if $y$ is $x$’s grandparent.
    \item \texttt{IsUncle}: \texttt{IsUncle}$(x, y)$ is \textit{True} if $y$ is $x$’s uncle.
    \item \texttt{IsMGUncle}: \texttt{IsMGUncle}$(x, y)$ is \textit{True} if $y$ is $x$’s maternal great uncle.
\end{itemize}


\subsubsection{Relations among different tasks}
We would like to clarify that these different tasks are indeed standard benchmarks for Inductive Logic Programming (ILP) [3,4,5]. They are commonly used in previous ILP works to evaluate different aspects (such as compositional relation prediction or property prediction) of the logic reasoning capability. First, each task includes its own task-dependent knowledge besides carrying common knowledge. Second, although each category of tasks (e.g., graph tasks and family tree tasks) shares a certain amount of common background knowledge, the objectives of the tasks (even in the same category) differ a lot due to each task’s individual information and goal (see our example below). For example, \texttt{1-OutDegree}, \texttt{AdjacentToRed}, and \texttt{4-Connectivity} are graph tasks and share the same background knowledge of \texttt{HasEdge}, which provides the information on the connectedness between two nodes in a graph. However, the goal of \texttt{1-OutDegree} is to predict if the out-degree of a node is exactly $1$, while that of \texttt{4-Connectivity} is to predict if the number of hops between two nodes is within $4$. \texttt{AdjacentToRed} has task-specific information on the color of each node (\texttt{IsRed}), and its goal is to predict if a node has an edge with a red node.

For convenience, we list an example below that compares between \texttt{AdjacentToRed}, \texttt{4-Connectivity}, and \texttt{1-OutDegree}. We can see that the corresponding logic expressions differ significantly, even if the graph topologies are the same.
The logic rule for \texttt{AdjacentToRed} is:
$$
\forall y \texttt{PredRed}(x, y) \leftarrow \texttt{IsRed}(x)
$$
$$
\texttt{AdjacentToRed}(x) \leftarrow \exists y (\texttt{HasEdge}(x, y) \wedge \texttt{PredRed}(x, y) ) ~.
$$
The logic rule for \texttt{4-Connectivity} is:
$$
\texttt{Distance2}(x, y) \leftarrow \exists z (\texttt{HasEdge}(x, z) \wedge \texttt{HasEdge}(z, y)) 
$$
$$
\texttt{Distance3}(x, y) \leftarrow \exists z (\texttt{Distance2}(x, z) \wedge \texttt{HasEdge}(z, y)) 
$$
$$
\texttt{Distance4}(x, y) \leftarrow \exists z (\texttt{Distance2}(x, z) \wedge \texttt{Distance2}(z, y)) 
$$
$$
\texttt{4-Connectivity}(x, y) \leftarrow \texttt{HasEdge}(x, y) \vee
           \texttt{Distance2}(x, y) \vee \texttt{Distance3}(x, y) \vee \texttt{Distance4}(x, y) ~.
$$
The logic rule for \texttt{1-OutDegree} is:
$$
\texttt{1-Outdegree}(x) \leftarrow \exists y \forall z (\texttt{HasEdge}(x, y) \wedge 	\neg \texttt{HasEdge}(x, z))~.
$$


\subsection{Hyper-parameter Settings}
\label{appendix:experiment_detail:hyper}
For simplicity, we set discount factor $\gamma = 1$ in all experiments. The $T_{\max}$ depends on the depth of the Reasoner.
In single-task setting, we use the same hyper-parameter settings as in the original papers for MemNN~\citep{sukhbaatar2015end}, $\partial$ILP~\citep{evans2018learning}, NLM~\citep{nlm}, and DLM~\citep{matthieu2021differentiable}.
That distribution for DLM-MTR is [$0.1, 0.12, 0.12, 0.1, 0.1, 0.13, 0.13, 0.2$]. 
The details of hyper-parameters in DLM-MTR can be found in Table~\ref{tab:params_mtr}.
Similarly, the ``Internal Logic'', initial temperature and scale of the Gumbel distribution, and the dropout probability in DLM-MTR are also kept the same settings as those in DLM. 
Besides, the probability distribution of sampling a task for PRIMA is [$0.1, 0.12, 0.12, 0.1, 0.1, 0.13, 0.13, 0.2$].
Regarding the hyper-parameters of PRIMA, details can be found in Table~\ref{table:params_pr}. We also show the hyper-parameters of PRISA-REINFORCE, PRISA-PPO, PRISA-MuZero in Table~\ref{table:params_pr}.

We note that the problem size for training is always $10$ for graph tasks and $20$ for family tree tasks in single-task and multi-task settings.


\begin{table}[]
    \centering
    \caption{Hyper-parameter settings for DLM--MTR.}
    \label{tab:params_mtr}
    \begin{tabular}{l c c}
        \toprule
         & DLM--MTR\\
         \midrule
         learning rate &  $0.005$ \\
         epochs &  $200$\\
         epoch size & $2000$\\
         batch size (train) & $4$\\
         breadth &$3$ \\
         depth &$9$ \\
         \bottomrule
    \end{tabular}
\end{table}

\begin{table}[]\caption{Hyper-parameter settings for PRISA and PRIMA. ``lr-reasoner'' refers to the learning rate for the reasoner, ``lr-policy'' denotes the learning rate for the policy network, ``lr-value'' denotes the learning rate for the value network,
``residual'' refers to the residual connection in the reasoner,
``NumWarmups'' refers to the number of warm-ups before starting the training,
``NumRollouts'' refers to the number of roll-outs in MCTS,
``RwdDecay'' refers to the constant exponential decay applied on the reward,
``$c_1$ and $c_2$'' refers to the same constants in PUCT formula in Section~\ref{appendix:algorithm_details:mcts},
``RBsize'' refers to the replay buffer size that is used to count the number of stored trajectories,
``BatchSize'' refers to the batch size for training,
``TrainingSteps'' refers to the number of training steps. For PRISA-REINFORCE, PRISA-PPO, PRISA-MuZero, the ``TrainingSteps'' denotes the training steps for each task, while that of PRIMA represents the training steps for all 8 tasks.
}\label{table:params_pr}
	\centering
	\begin{tabular}{ l c  c  c  c}
	    \toprule
		& PRISA- & PRISA- & PRISA- & PRIMA\\
		& REINFORCE & PPO & MuZero  &  \\
		\midrule
		lr-reasoner  & $0.005$ & $0.005$ & $0.005$ & $0.004$ \\
		lr-policy  &$0.085$ & $0.085$ & $0.075$ & $0.075$ \\
		lr-value  &- & $0.15$ & $0.075$ & $0.075$ \\
		breadth  &$3$ & $3$ & $3$ & $3$ \\
		depth  &$4$ & $4$ & $4$ & $4$ \\
		residual & \textit{False} & \textit{False} & \textit{False} & \textit{False} \\
		NumWarmups  & - & - & $200$ & $200$ \\
		NumRollouts  &- & - & $1200$ & $1200$ \\
		RwdDecay  &$5$ & $5$ & $5$ & $5$ \\
		$c_1$  & - & - & $30$ & $30$\\
		$c_2$  & - & - & $19652$ & $19652$\\
		RBsize  & $16$ & $32$ & $400$ &$400$ \\
		BatchSize  & $16$ & $16$ & $16$ &$16$ \\
		TrainingSteps  & $7\times10^{4}$ & $7\times10^{4}$ & $7\times10^{4}$ & $39\times10^{4}$ \\
		\bottomrule 
	\end{tabular}
\end{table}

\subsection{Computing infrastructure}
\label{appendix:experiment_detail:infrastructure}
We conducted our experiments on a CPU server where the CPU is ``Intel(R) Xeon(R) Silver 4114 CPU'' with $40$ cores and $64$ GB memory in total.


\begin{table}[]
\centering
\caption{Additional results: The performance of PRIMA w.r.t different numbers of intermediate predicates. ``\# pred'': the number of intermediate predicates.}
\label{tab:prima-dim}
\resizebox{\textwidth}{!}{%
\begin{tabular}{|c|l|l|lllll|l|lll|}
\hline
\multirow{13}{*}{\rotatebox[origin=c]{90}{testing accuracy}} &                                                  & Family Tree & HasFather & HasSister & IsGrandparent & IsUncle & IsMGUncle & Graph & AdjacentToRed & 4-Connectivity & 1-OutDegree \\ \cline{2-12} 
                                   & \multirow{3}{*}{\# pred = 2}                 & m=20        & 61.78     & 54.97     & 96.46         & 97.19   & 99.67     & m=10  & 42.62         & 89.28          & 81.88       \\
                                   &                                                  & m=100       & 87.5      & 31.2      & 97.7          & 96.5    & 98.4      & m=50  & 11.1          & 96.6           & 96.5        \\
                                   &                                                  & PSS         & 0         & 0         & 0             & 0       & 0         & PSS   & 0             & 0              & 0           \\ \cline{2-12} 
                                   & \multirow{3}{*}{\# pred = 4}                 & m=20        & 100       & 100       & 100           & 97.2    & 99.7      & m=10  & 67.3          & 93.4           & 100         \\
                                   &                                                  & m=100       & 100       & 100       & 100           & 96.6    & 98.4      & m=50  & 91.7          & 97.8           & 100         \\
                                   &                                                  & PSS         & 100       & 100       & 100           & 0       & 0         & PSS   & 0             & 0              & 100         \\ \cline{2-12} 
                                   & \multirow{3}{*}{\# pred = 6}                 & m=20        & 100       & 100       & 100           & 100     & 100       & m=10  & 99.7          & 100            & 100         \\
                                   &                                                  & m=100       & 100       & 100       & 100           & 100     & 100       & m=50  & 100           & 100            & 100         \\
                                   &                                                  & PSS         & 100       & 100       & 100           & 100     & 90        & PSS   & 0             & 100            & 100         \\ \cline{2-12} 
                                   & \multirow{3}{*}{\# pred = 8} & m=20        & 100       & 100       & 100           & 100     & 100       & m=10  & 100           & 100            & 100         \\
                                   &                                                  & m=100       & 100       & 100       & 100           & 100     & 100       & m=50  & 100           & 100            & 100         \\
                                   &                                                  & PSS         & 100       & 100       & 100           & 100     & 90        & PSS   & 90            & 100            & 100         \\ \hline
\end{tabular}%
}
\end{table}

\section{Additional Experiments}
\label{sec:appendix:add-exp}

\subsection{Reasoning costs of PRISA}
\begin{figure*}
    \centering
    \includegraphics[width=.98\textwidth]{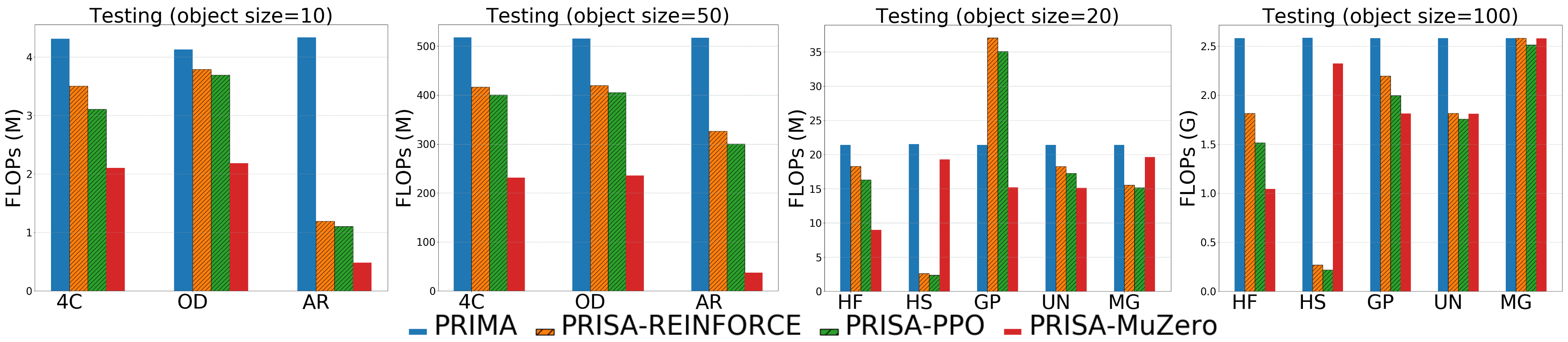}
    \caption{\footnotesize{The reasoning costs (in FLOPs) of PRIMA and three variants of PRISA at the inference stage (1 M=1$\times 10^6$, 1 G=1$\times 10^9$).}}
    \label{fig:comp-flops2}
\end{figure*}

To demonstrate the improved reasoning efficiency in PRISA at the inference stage, three variants of PRISA (e.g., PRISA-REINFORCE, PRISA-PPO, and PRISA-MuZero) are compared with PRIMA in terms of FLOPs. 
As shown in Fig.~\ref{fig:comp-flops2}, three variants of PRISA generally have lower reasoning complexity than PRIMA. 
The difference in FLOPs between PRIMA and PRISA-MuZero is primarily from the difference between multi-tasking settings and single-task settings.

\subsection{Predicate dimension analysis}
Since the Reasoner only realizes a partial set of Horn clauses of FOL, the number of intermediate predicates will greatly determine the expressive power of the model. The performance of PRIMA degrades gracefully when the number of intermediate predicates decreases. 
This is shown in the additional experiments of examining the limiting performance of our PRIMA, where we examine the limiting performance of our PRIMA when the number of intermediate predicates at each layer decreases.
The results in Table~\ref{tab:prima-dim} show that the performance of PRIMA degrades gracefully when the number of intermediate predicates decreases; for example, it still performs reasonably well on most tasks even when the number of predicates becomes 4.

\subsection{Analysis of multi-tasking capabilities}
%
\begin{figure*}[htb!]
    \centering
    \includegraphics[width=.90\textwidth]{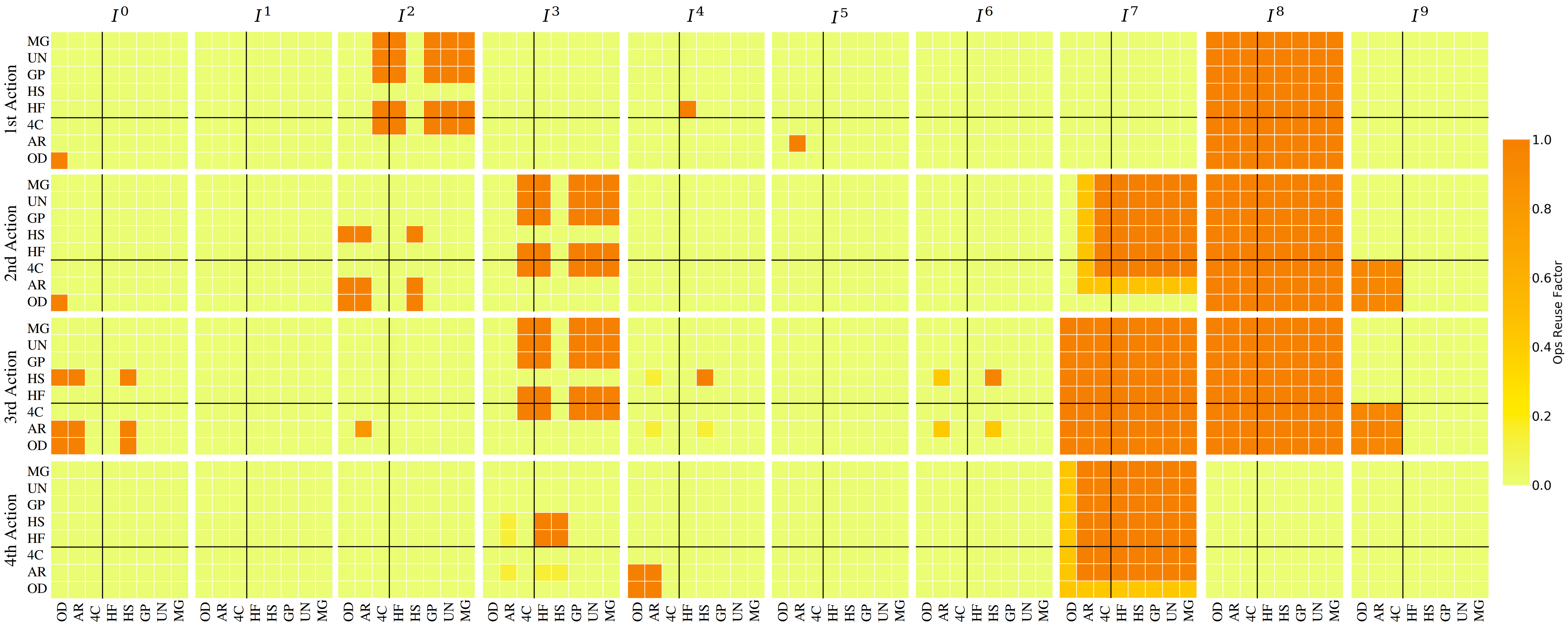}
    \caption{\footnotesize{The quantitative evaluation of ops reuse between different Tasks in PRIMA.}}
    \label{fig:htmp-ovl}
\end{figure*}

\vspace{-.1cm}

\paragraph{Why/how do different tasks share blocks (of ops) as in Fig.~\ref{fig:htmp-ovl}?}
To quantitatively measure how different tasks reuses the ``skills'', we show in Fig.~\ref{fig:htmp-ovl}, which is a heat-map showing the probability of overlapping about Ops reuse across different tasks, where red color denotes a high-level of Ops overlapping.
Ops reuse factor is defined as $\min(P_a(I^j), P_b(I^j))$ where $P_a(I^j)$ refers to the probability of ops $I^j (j \in \{0, \ldots, 9\})$ being active for task $a$ and $P_b(I^j)$ denotes the probability of $I^j$ being active for task $b$.
From Fig.~\ref{fig:htmp-ovl}, it can be inferred that some tasks share skills more than others. For example, in most of the actions, $I^8$ is shared across different tasks.

\paragraph{Does the $100\%$ accuracy imply these testbeds are too simple?}
In fact, these tasks are quite challenging for deep learning (DL) approaches (e.g., the poor performance of MemNN in Table~\ref{tab:x6}). In particular, they are designed to evaluate the model’s out-of-distribution generalization capabilities. For example, in Table~\ref{tab:x6}, when the deep learning model (MemNN) is tested at a problem size (m=100 in Family tree tasks) that has never been seen during training (i.e., MemNN is trained on Family tree tasks with m=20), its performance will degrade drastically. It shows that DL-based approaches cannot learn the correct logical rules for solving these problems. In contrast, logic reasoning methods (e.g., ILP/ NLM/ DLM) can (and are expected to) reach 100\% generalization accuracy on those tasks. In other words, reaching 100\% accuracy is a “must-be” criterion for those logic reasoning methods. For this reason, Tables~\ref{tab:x5}\&~\ref{tab:x6} are not used to show our improvement in accuracy but to justify that PRIMA can meet the expectation of perfect generalization (aka, 100\% accuracy). Under such a “perfect generalization” constraint, our major objective is to improve the reasoning efficiency (i.e., reducing FLOPs) over other approaches.

\subsection{The significance of reducing FLOPs in practice}
First, the comparison between PRIMA and the baseline (e.g., NLM) is indeed fair in that the expressiveness of NLM and that of the Reasoner module in PRIMA are similar (i.e., same hidden dimensions and number of layers --- see Appendix D of the supplementary material). So NLM is not using more FLOPs for expressive power. Note that the difference lies in the Planner module of PRIMA, which NLM doesn’t have. We would also like to highlight that reducing the computational cost (in FLOPs) of reasoning has significant meanings in practice.

\begin{itemize}
    \item For example, when we deploy a trained model as a cloud API service, the model will receive a massive amount of API calls (e.g., more than 1 million API calls per day for speech/image recognition service), which accounts for a huge number of repeated model inference operations on GPU servers. Although cutting the reasoning time by half may not seem much for one single input sample, repeating it a massive amount of times during inference over months or years will incur a huge cost. (The aggregated inference cost is even significantly higher than the training cost, where training is generally incurred every 6-12 months in such systems.) In the multi-task reasoning setting, we believe this behavior will be even more outstanding as a single model is deployed to handle many different types of services and tasks. When translated into spending cost, if we can reduce the FLOPs by half, we can generally cut the spending cost on cloud GPU servers by half, which would be a significant cut for operating spending to host these cloud API services.
    
    \item Another important application where reducing FLOPs is essential is when the trained model is deployed on mobile/edge devices. In this situation, the computation task of the model must be accomplished with limited resource supplies, such as computing time, storage space, battery power, etc.
    
    \item In addition, we would also like to point out that improving the inference efficiency of general machine learning models has attracted a significant amount of interest from both academia and industry.
\end{itemize}

\subsection{Training cost for different MTR models}
We here report the approximate training costs of different methods below (in total training time) in order to give readers an overall picture of the overhead (running on CPU clusters):
\begin{itemize}
    \item DLM-MTR = 4$\sim$5days
    \item PRIMA = 13$\sim$14 days
\end{itemize}

In summary, PRIMA tends to have a higher training cost: about 2.6$\sim$3.5 times that of DLM-MTR. This is not surprising since multiple rollouts of MCTS would generate additional overheads. Nevertheless, it should be noted that such an overhead is still affordable and could be amortized if the inference efficiency is more important as in the cloud API service.

\section{Additional Related Work}
\label{sec:appendix:related}
This section provides complementary contents to Sec.~\ref{sec:related}.

\paragraph{Model-based symbolic planning}
Another large body of related work is symbolic planning (SP)~\citep{van2008handbook,hanheide2015robot,chen2016planning,khandelwal2017bwibots}. In SP approaches, a planning agent carries prior symbolic knowledge of objects, properties, and how they are changed by executing actions in the dynamic system, represented in a formal, logic-based language such as PDDL \citep{mcdermott1998pddl} or an action language~\citep{gel98} that relates to logic programming under answer set semantics (answer set programming)~\citep{lif08}. 
The agent utilizes a symbolic planner, such as a PDDL planner \textsc{FastDownward}~\citep{helmert2006fast} or an answer set solver \textsc{Clingo}~\citep{gekasc12c} to generate a sequence of actions based on its symbolic knowledge, executes the actions to achieve its goal. Compared with RL approaches, an SP agent does not require a large number of trial-and-error to behave reasonably well, yet requires predefined symbolic knowledge as the model prior. 
It should be noted that it remains questionable if PDDL-based methods can be applied directly to our problem, as there is no explicitly predefined prior knowledge under our problem setting, which is required by PDDL-based approaches. In contrast, we address the learning-to-reason problem via a complete data-driven deep reinforcement approach.

\paragraph{Multi-task RL}
Multi-task learning can help boost the performance of reinforcement learning, leading to multi-task reinforcement learning~\citep{mtrl:vithayathil2020survey,mtrl:zhu2020transfer}. Some research \citep{wfrt07,llc09b,lg10,clr14,akl17,dds17,ser17,bbrw19,vnnbkttl19,ignsbw20} has adapted the ideas behind multi-task learning to RL.
For example, in \citep{bbrw19}, a multi-task deep RL model based on attention can group tasks into sub-networks with state-level granularity.
The idea of compression and distillation has been incorporated into multi-task RL as in \citep{pbs16,rcgdkpmkh16,opahv17,tbcqkhhp17}.
For example, in \citep{pbs16}, the proposed actor-mimic method combines deep reinforcement learning with model-compression techniques to train a policy network that can learn to act for multiple tasks.
Other research in multi-task RL focuses on online and distributed settings \citep{bbt17,sr17,sjhr18,esmsmwdf18,tkb18,lbkh19}.

\paragraph{RL for logic reasoning}
Relational RL integrates RL with statistical relational learning and connects RL with classical AI for knowledge representation and reasoning. Most prominently, \citet{dvzeroski2001relational} originally proposed relational RL, \citet{tadepalli2004relational} surveyed relational RL, \citet{guestrin2003generalizing} introduced
relational MDPs, and \citet{diuk2008object} introduced objected-oriented MDPs (OO-MDPs). More recently, \citet{battaglia2018relational} proposed to incorporate relational inductive bias, \citet{zambaldi2018relational}
proposed deep relational RL, \citet{keramati2018strategic} proposed strategic object-oriented RL, and some researchers have also adopted deep learning approaches for dealing with relations and/or reasoning, for example \citep{battaglia2016interaction,chen2018iterative,santoro2017simple,santurkar2018does,palm2017recurrent,xia2018learning,yi2018neural}.


\paragraph{Statistical relational reasoning}
Statistical relational reasoning (SRL)~\citep{koller2007introduction} often provides a better understanding of domains and predictive accuracy, but with more complex learning and inference processes. Unlike ILP, SRL takes a statistical and probabilistic learning perspective and extends the probabilistic formalism with relational aspects. Examples of SRL include text classification \citep{ganiz2010higher}, recommendation systems~\citep{cao2016non}, and wireless networks \citep{yoshida2020hybrid}.
SRL can be divided into two research branches: probabilistic relational models (PRMs) \citep{koller1999probabilistic} and probabilistic logic models (PLMs)~\citep{chen2008learning}. PRMs start from probabilistic graphical models and then extend to relational aspects, while PLMs start from ILP and extend to probabilistic semantics.
A related research area is graph relational reasoning, such as using graph neural networks~\citep{zhou2020graph} to conduct reasoning. 
A discussion of this approach is beyond the scope of this paper, but we refer interested readers to~\citep{chakrabarti2006graph,cook2006mining,zhang2019graph} for a comprehensive review.

\paragraph{Comparisons with NLM}
\label{sec:appendix:related:nlm}
The Reasoner shares the same functionality with NLM, in terms of approximating the meta-rules ($\mathtt{Boolean Logic}$, $\mathtt{Expansion}$, and $\mathtt{Reduction}$).
However, our Reasoner has major differences from NLM.
\textit{First}, the targets are different. Our Reasoner targets the trade-off between capability and efficiency. Compared with NLM, our Reasoner can greatly improve efficiency by using a Planner module.
\textit{Second}, our Reasoner has more explicitness in the reasoning path.
Let $[O^{r-1}_{i-1}, O^{r}_{i-1}, O^{r+1}_{i-1}]$ denote the output predicates (in tensor representation) from the previous layer, and $\mathrm{Concate}(\cdot)$ denotes the concatenation operation. NLM performs the intra-group computation as
\begin{align}
O_{i}^{r} = \sigma\left(
  \mathrm{MLP}\left(\mathrm{Permute}\left(Z_i^{r}\right); \theta_i^{r}\right) \right),
\end{align}
  where $O_i^{r}$ is the output predicate, $Z_{i}^{r} = \mathrm{Concat}\left(
      \mathrm{Expand}\left(O_{i-1}^{r-1}\right),
      O_{i-1}^{r},
      \mathrm{Reduce}\left(O_{i-1}^{r+1}\right)
\right)$, $\sigma$
is the sigmoid nonlinearity and $\theta_i^{r}$ denotes learnable parameters. 
On the contrary, in the Reasoner module of our PRISA/PRIMA, the implementation is
\begin{align}
    O_{i}^{r} = \sigma \Bigg(
  \mathrm{MLP}(\mathrm{Permute}\big(\mathrm{Expand}(O_{i-1}^{r-1})\big); \theta_i^{r}) + \mathrm{MLP}(\mathrm{Permute}\big(O_{i-1}^{r}); \theta_i^{r}\big) \\
  + \mathrm{MLP}(\mathrm{Permute}\big(\mathrm{Reduce}(O_{i-1}^{r+1})); \theta_i^{r} \big) \Bigg),
\end{align}
which allows for the neural implementations of $\mathtt{Boolean Logic}$, $\mathtt{Expansion}$, and $\mathtt{Reduction}$ (at the same arity group) to be executed independently to some extent and reduce the unnecessary computations.

\paragraph{Comparisons with chaining in logic reasoning}
Our main focus is to develop a neural architecture that learns to perform efficient (first-order logic) reasoning. There are two types of chaining to conduct logic reasoning, i.e., forward-chaining and backward-chaining. The forward-chaining strategy deduces all the possible conclusions based on (all) the available facts and deduction rules until it reaches the desired conclusion. (The other strategy is backward-chaining, which starts from the desired conclusion (i.e., the goal) and then works backward recursively to validate the available facts.) This paper adopts the forward-chaining approach (of the learnable Horn clauses) to perform logic reasoning. Note that it is generally unrealistic to generate all the possible conclusions based on all the available deduction rules as in the vanilla forward-chaining strategy. Therefore, our focus in the paper is to learn a neural module (named “planner”) capable of identifying a small subset of paths in the forward-chaining strategy that can lead to the correct solution (as well as learning the Horn clauses in reasoner). 
If the “planner” module does not chain the correct subset of Horn clauses (logic operators), it will obviously NOT give us the correct solution. Therefore, we need to select a subset of paths that can guarantee the correct answer with low cost, which is reflected in our RL reward design as well (see Eq.~(\ref{eq:r}) in the paper).

\paragraph{Comparisons with~\citet{minton1990quantitative}}
There are several major differences between our work and ~\cite{minton1990quantitative} regarding motivations, methods, and technical details. We list a few here.

First, both PRIMA and~\cite{minton1990quantitative} can learn to improve reasoning regarding the utility issue, while PRIMA uses RL to achieve this goal.

Second, PRIMA differs from~\cite{minton1990quantitative} in several perspectives, including input information/problem settings. EBL requires that the input domain knowledge must be formulated via rule-based logic. In contrast, the domain knowledge can be probabilistic transition functions in PRIMA, depending on the concrete MBRL method adopted. EBL follows an explain-generalize-refine procedure (detailed see below). Our PRIMA follows another style: we first learn the neural operators equivalent to (in functionality) the explainable logic operations and then learn the solution paths expanded based on these neural operators.
multi-task capability. PRIMA is designed for multi-task reasoning, while it remains unclear in~\cite{minton1990quantitative} if EBL can conduct multi-task reasoning and under what scenarios.
EBL features the following steps:
\begin{enumerate}
    \item Explain: explaining the observed target value for this example in terms of the domain theory.
    \item Generalize: analyzing this explanation to determine the general conditions under which the explanation holds.
    \item 
Refine: refining its hypothesis to incorporate these general conditions.
\end{enumerate}